\documentclass[preprint,12pt,number,sort&compress]{elsarticle}

\usepackage{graphicx}
\usepackage{amsmath, amssymb}
\usepackage{hyperref}
\usepackage{caption}
\usepackage{subcaption}
\usepackage{multirow}
\usepackage{color}
\usepackage{booktabs}
\usepackage{svg}

\journal{Engineering Applications of Artificial Intelligence}

\begin{document}

\begin{frontmatter}

\title{YOLO-SPCI: Enhancing Remote Sensing Object Detection via Selective-Perspective-Class Integration}

\author[1]{Xinyuan Wang}
\author[2]{Lian Peng}
\author[1]{Xiangcheng Li}
\author[3]{Yilin He}
\author[1]{\texorpdfstring{KinTak U\corref{cor1}}{KinTak~U}}

\address[1]{Faculty of Innovation Engineering, Macau University of Science and Technology, Macau, China}
\address[2]{School of Computer and Information Science, Southwest University, Chongqing, China}
\address[3]{College of Automation and Information Engineering, Nanjing University of Posts and Telecommunications, Nanjing, China}

\cortext[cor1]{Corresponding author. Email: ktu@must.edu.mo}

\begin{abstract}
Object detection in remote sensing imagery remains a challenging task due to extreme scale variation, dense object distributions, and cluttered backgrounds. While recent detectors such as YOLOv8 have shown promising results, their backbone architectures lack explicit mechanisms to guide multi-scale feature refinement, limiting performance on high-resolution aerial data. In this work, we propose YOLO-SPCI, an attention-enhanced detection framework that introduces a lightweight Selective-Perspective-Class Integration (SPCI) module to improve feature representation. The SPCI module integrates three components: a Selective Stream Gate (SSG) for adaptive regulation of global feature flow, a Perspective Fusion Module (PFM) for context-aware multi-scale integration, and a Class Discrimination Module (CDM) to enhance inter-class separability. We embed two SPCI blocks into the P3 and P5 stages of the YOLOv8 backbone, enabling effective refinement while preserving compatibility with the original neck and head. Experiments on the NWPU VHR-10 dataset demonstrate that YOLO-SPCI achieves superior performance compared to state-of-the-art detectors.
\end{abstract}

\begin{keyword}
Remote Sensing \sep Object Detection \sep Multi-Dimensional Attention \sep YOLO \sep Lightweight Backbone
\end{keyword}

\end{frontmatter}

\section{Introduction}

Remote sensing object detection is a fundamental task with broad applications in urban planning, environmental monitoring, and military surveillance~\cite{Li2024Military, Zhang2022OCOD, weng2024enhancing, qiao2022novel, Li2024FewShot, Yue2024HRNet, AlGaradi2024UAV, Li2023SCFNet, Liu2024YOLOv5Ship}. The increasing availability of high-resolution aerial imagery provides rich visual data but also introduces significant challenges: extreme object scale variation, dense spatial distributions, and cluttered backgrounds demand detectors with robust and discriminative multi-scale representations.

Recent advances such as YOLOv8~\cite{yolov8}, Deformable DETR~\cite{zhu2020deformable}, and PR-Deformable DETR~\cite{chen2024pr} have demonstrated strong performance through the use of advanced backbones and refined detection heads. These models often incorporate attention mechanisms to enhance feature quality. For instance, Squeeze-and-Excitation (SE)~\cite{hu2018squeeze} adaptively reweights channel-wise features, while CBAM~\cite{woo2018cbam} extends this by introducing both channel and spatial attention. However, directly applying such modules to remote sensing imagery remains suboptimal due to three main limitations: first, most attention mechanisms operate along a single dimension, limiting their ability to integrate diverse contextual cues; second, local dependency modeling is often emphasized at the expense of global context or class-aware relevance, both critical for dense aerial scenes; and third, existing modules designed for natural image benchmarks may incur prohibitive computational costs when scaled to high-resolution remote sensing inputs.

To address the above challenges in small object detection, context modeling, and semantic confusion in remote sensing imagery, we propose a novel detection framework termed {YOLO-SPCI}, which augments the YOLOv8 backbone with a lightweight and generalizable attention module called {Selective-Perspective-Class Integration (SPCI)}. Specifically, the SPCI module enhances feature representations from three complementary dimensions: (1) a {Selective Stream Gate (SSG)} for capturing long-range dependencies and adaptively controlling information flow; (2) a {Perspective Fusion Module (PFM)} for aggregating multi-path contextual features; and (3) a {Class Discrimination Module (CDM)} for refining class-aware structures to improve semantic separability. These three components are integrated into a unified block that can be flexibly embedded into standard backbones.
We instantiate two SPCI modules at stages P3 and P5 of the YOLOv8 backbone, targeting both shallow high-resolution and deep semantic layers. This design enables joint improvement of small object localization and large object recognition, while keeping the neck and detection head untouched to ensure compatibility and fair comparison.
\vspace{0.5em}
\noindent{Our main contributions are summarized as follows:}
\begin{itemize}
    \item We propose a novel {Selective-Perspective-Class Integration (SPCI)} module that unifies global context modeling, multi-path fusion, and class-discriminative refinement into a lightweight attention unit.
    \item We integrate SPCI into the YOLOv8 backbone at both low-level and high-level stages, improving feature quality across scales while maintaining architectural simplicity and high efficiency.
    \item We validate the proposed YOLO-SPCI on the NWPU VHR-10 dataset, where it consistently outperforms state-of-the-art baselines in both accuracy and speed, demonstrating its effectiveness and generalization in challenging remote sensing detection scenarios.
\end{itemize}

\section{Related work}

First, we provide an overview of recent advancements in object detection, including multiscale detection methods, attention mechanisms, and their applications in remote sensing scenarios.

\subsection{Object detection}

Object detection remains a core problem in computer vision, requiring the accurate localization and classification of objects within complex scenes. Over the past decade, detection pipelines have evolved from early handcrafted feature-based methods to deep learning-based frameworks that now dominate the field. Two-stage methods such as Faster R-CNN~\cite{ren2015faster} first generate object proposals and then refine them, offering high accuracy at the cost of computational speed. In contrast, one-stage detectors like SSD~\cite{liu2016ssd} and YOLOv1~\cite{redmon2016you} predict bounding boxes and class scores in a single forward pass, dramatically improving inference speed. Among these, the YOLO series~\cite{redmon2018yolov3, bochkovskiy2020yolov4, wang2020cspnet, yolov5github, yolov8} has gained widespread adoption due to its effective balance between speed and precision. YOLOv8 in particular brings notable improvements in architecture modularity, scalability, and training efficiency, making it a popular backbone for downstream enhancements.

Despite the progress, existing YOLO-based models still struggle with small object detection, multi-scale localization, and robustness under cluttered backgrounds. These limitations have prompted growing interest in augmenting detection backbones with specialized modules for feature refinement. Inspired by recent advances in modular visual architectures~\cite{shen2025imaggarment, shen2025long}, we propose the Selective-Perspective-Class Integration (SPCI) module, which unifies three components: a Selective Stream Gate (SSG) for dynamic global flow regulation, a Perspective Fusion Module (PFM) for context-aware aggregation, and a Class Discrimination Module (CDM) for enhancing semantic separability. Embedded within the YOLOv8 backbone, SPCI strengthens its multi-scale representations without compromising computational efficiency, leading to improved detection performance across object scales and categories.

\subsection{Challenges in remote sensing object detection}
Remote sensing imagery, known for its ultra-high resolution and wide spatial coverage, poses distinct challenges for object detection. Targets in these images are often small, densely packed, and situated within highly cluttered backgrounds. In addition, remote sensing scenes exhibit extreme variations in object scale and uneven spatial distributions, making multi-scale detection particularly difficult. The high cost of manual annotation and severe class imbalance further hinder model generalization and robustness.
To mitigate these issues, prior works have proposed a variety of architectural and algorithmic strategies. Multi-scale feature aggregation methods such as Feature Pyramid Networks (FPN)~\cite{lin2017feature} and High-Resolution Networks (HRNet)~\cite{wang2020deep} enhance small object detection by preserving resolution across layers. Attention mechanisms like SE~\cite{hu2018squeeze} and CBAM~\cite{woo2018cbam} help suppress background noise and focus feature responses on relevant regions. 

Lightweight backbones such as MobileNet~\cite{howard2017mobilenets} have also been explored to reduce inference latency in high-resolution scenarios. Despite these advances, current approaches still fall short in balancing precision, efficiency, and robustness—particularly when dealing with small objects and fine-grained semantics in remote sensing images.
Motivated by recent advances in unified generative modeling frameworks~\cite{shen2024imagpose, shen2025imagdressing}, we introduce a task-specific attention module named \textbf{SPCI}, which integrates three key components: channel-wise selection, spatial-context fusion, and class-aware semantic enhancement. Designed to address the specific demands of remote sensing detection, SPCI improves feature quality for small targets, enhances spatial resolution fusion, and strengthens inter-class discrimination. This unified structure enables precise and robust object detection under complex conditions while preserving computational efficiency, making it highly suitable for high-resolution aerial image analysis.

\subsection{Attention mechanism}

Recent years witness extensive applications of attention mechanisms in object detection, with research primarily focusing on channel and spatial attention. Classic methods such as the SE module leverage global average pooling to dynamically adjust the weights of feature channels, significantly improving feature selection efficiency. The CBAM module further integrates spatial attention, dynamically assigning weights to salient regions, enabling models to capture critical features more precisely. These methods excel at optimizing local feature selection and spatial relationships but often overlook the modeling of class-specific characteristics.

In remote sensing object detection, methods like PR-Deformable DETR \cite{chen2024pr} effectively combine Deformable Attention \cite{zhu2020deformable} and Transformer-based global modeling capabilities \cite{carion2020end}. These approaches flexibly capture local features and model long-range dependencies, significantly enhancing small object detection performance. Particularly in multi-scale scenarios, PR-Deformable DETR dynamically focuses on key regions, demonstrating robust performance and offering valuable insights into feature modeling for remote sensing images.

Our works propose the SPCI module, which incorporates CDM to address the limitations of existing attention mechanisms in class-specific modeling. CDM dynamically optimizes class-specific feature representations, enhancing the model's ability to differentiate between classes in complex environments. By focusing on the deep modeling of class features, CDM provides robust support for object detection in remote sensing images, achieving a balance between lightweight design and high accuracy.

\section{Method}

\begin{figure*}[t]
    \centering
    \includegraphics[width=\textwidth]{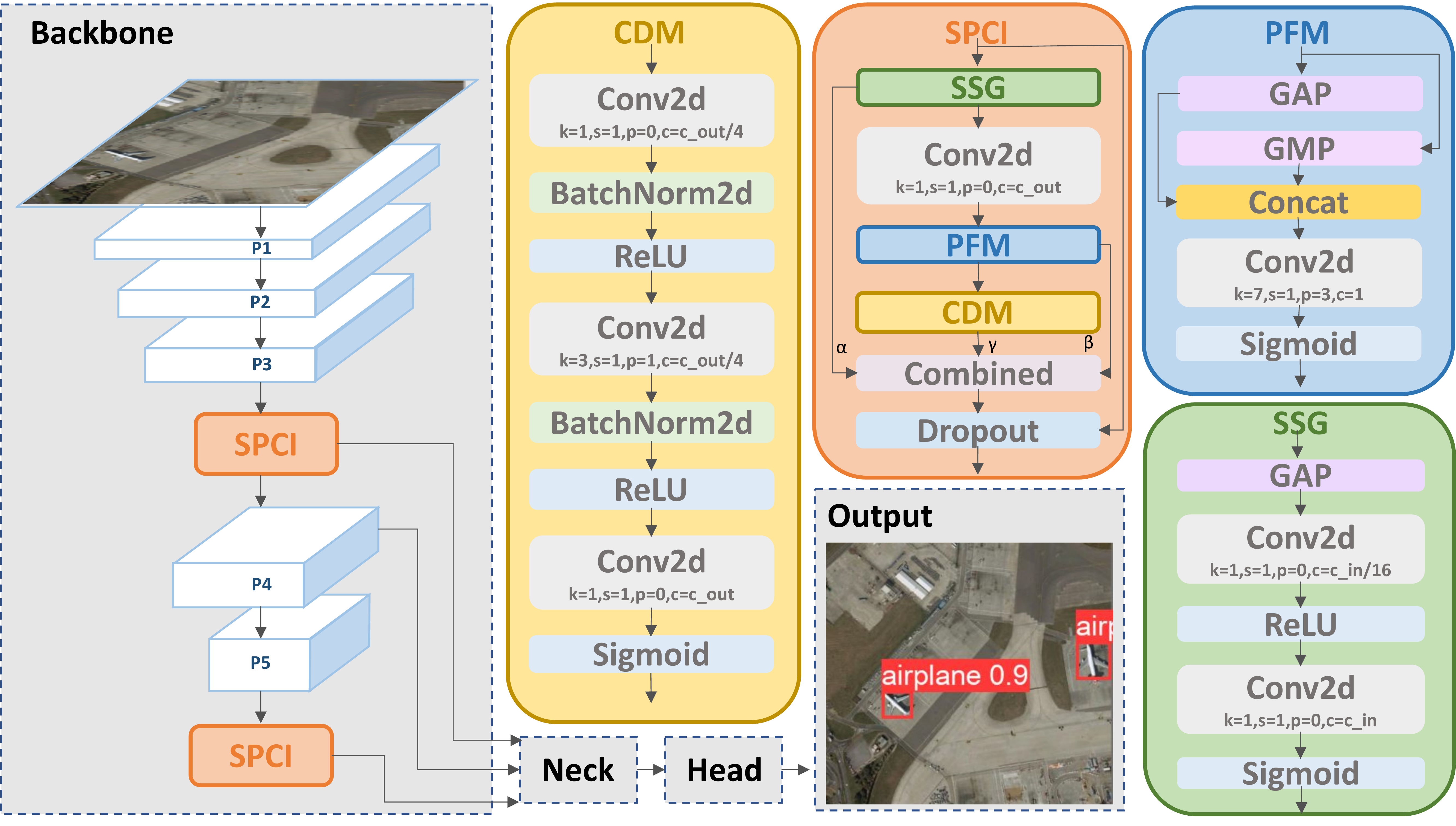} 
    \caption{The proposed selective-perspective-class integration (SPCI) module integrates the selective stream gate (SSG), perspective fusion module (PFM), and class discrimination module (CDM), and is embedded into the YOLOv8 backbone to enhance feature representation for object detection.}
    \label{fig:structure}
\end{figure*}

This section presents the overall architecture and core components of the proposed approach. We begin by describing the integration of our SPCI module into the YOLOv8 backbone, highlighting the motivation and placement strategy. Then, we elaborate on the internal design of SPCI, which consists of three specialized units: SSG, PFM, and CDM. Each unit is constructed to enhance feature representations from a complementary perspective, and their respective roles are detailed with structural explanations and mathematical formulations.

\subsection{Overall architecture}
In this work, we improve the YOLOv8 architecture by modifying its backbone through the insertion of a lightweight attention module, termed SPCI. The original neck and detection head are retained to maintain architectural consistency and ensure compatibility with the standard detection framework.
As shown in Fig.~\ref{fig:structure}, we embed the SPCI module after the P3 and P5 stages of the backbone. These two layers correspond to feature maps at different semantic levels and spatial resolutions: P3 provides higher-resolution representations that are better suited for locating small-scale targets, while P5 captures more abstract semantic features that contribute to the recognition of larger objects. Additionally, both P3 and P5 are directly connected to the neck for multi-scale feature fusion, making them important sources of information for the final detection process.
We introduce the SPCI module at these two points to enhance intermediate representations from multiple perspectives. The module integrates three subcomponents, each addressing a specific dimension of feature refinement. SSG dynamically adjusts the flow of feature information through global context modeling, allowing the network to emphasize relevant patterns while suppressing redundant ones. PFM introduces additional spatial structure by aggregating context features from average and max pooling branches, helping preserve structural cues across different object scales. CDM focuses on semantic-level refinement, strengthening the distinctiveness between object classes through local transformation and attention weighting. These components are structurally integrated to form the complete SPCI module, which is directly embedded into the backbone without altering the overall detection framework.

\subsection{Selective-Perspective-Class Integration Module}
The internal structure of the SPCI module is illustrated in Fig.~\ref{fig:structure}. It follows a sequential yet partially parallel design, where three functionally distinct submodules SSG, PFM, and CDM are integrated and connected through a combination of transformation, fusion, and residual connections.
The input feature first passes through SSG, which applies channel-wise attention based on global statistics. The output is then transformed via a $1 \times 1$ convolution to match the target output dimension. Subsequently, PFM is applied to incorporate spatial cues through pooled-context fusion, followed by CDM for enhancing semantic distinctiveness at the class level.
To preserve informative patterns from different stages and improve gradient flow, three outputs—denoted as $\alpha$ (from SSG), $\beta$ (from PFM), and $\gamma$ (from CDM)—are combined by element-wise addition before being passed to a dropout layer. This residual-style fusion enables the module to balance direct signals with progressively refined features.
The SPCI module is inserted after the P3 and P5 stages of the backbone. Its output, maintaining the same spatial size as the input, is forwarded to the neck for multi-scale feature fusion, allowing the enhanced representations to contribute directly to the detection head.

\subsubsection{Selective Stream Gate }
SSG is designed to modulate feature responses by leveraging global contextual information. As shown in Fig.~\ref{fig:structure}, it adopts a compact structure consisting of Global Average Pooling (GAP) followed by two point-wise convolutional layers.
Given an input feature map, SSG first applies GAP to extract a global descriptor summarizing the spatial content across all feature streams. This descriptor is then processed by a pair of $1 \times 1$ convolutions: the first reduces the number of feature streams to a compact intermediate size, followed by a ReLU activation, and the second restores it to the original dimension. A sigmoid function is applied to produce a set of weighting factors.
These factors are used to scale the original input feature map, selectively enhancing informative streams while suppressing less relevant ones. By doing so, SSG enables the network to regulate the internal flow of feature information with minimal computational cost, while preserving the spatial resolution and structural consistency of the input.
The weight vector $w_s$ is computed from the global descriptor of $F_s$, as shown in Equation~\eqref{eq:ssg_weight}. The enhanced output $F_s'$ is then obtained by element-wise multiplication between $F_s$ and $w_s$, as shown in Equation~\eqref{eq:ssg_output}:

\begin{equation}
w_s = \sigma \left( \text{Conv}_2 \left( \text{ReLU} \left( \text{Conv}_1 \left( \text{GAP}(F_s) \right) \right) \right) \right),
\label{eq:ssg_weight}
\end{equation}

\begin{equation}
F_s' = F_s \times w_s.
\label{eq:ssg_output}
\end{equation}

Here, $F_s$ denotes the input to the SSG module, and $F_s'$ is the corresponding output. $\text{GAP}(\cdot)$ represents global average pooling, $\text{Conv}_1$ and $\text{Conv}_2$ are 1$\times$1 convolutional layers, and $\sigma(\cdot)$ is the sigmoid activation. The weighting vector $w_s$ is applied to $F_s$ by element-wise multiplication across the stream dimension.

\subsubsection{Perspective Fusion Module }

PFM is designed to refine feature representations by integrating both global and local information. As shown in Fig.~\ref{fig:structure}, PFM uses GAP and Global Max Pooling (GMP) to capture different perspectives of the input feature map. The results from these operations are concatenated and passed through a $7 \times 7$ convolutional layer to enhance the feature map, followed by a sigmoid activation to produce the attention weights.

For an input feature map, PFM first applies GAP and GMP to extract global descriptors. GAP aggregates the overall context by averaging all values across the feature map, while GMP selects the most prominent features. These two descriptors are concatenated along the channel dimension and then processed by a convolutional layer. The weight vector $w_p$ is computed as follows, as shown in Equation~\eqref{eq:pfm_weight}.The enhanced output $F_p'$ is calculated by element-wise multiplication between the input feature map $F_p$ and the learned weight vector $w_p$, as shown in Equation~\eqref{eq:pfm_output}:

\begin{equation}
w_p = \sigma \left( \text{Conv}_1 \left( \text{Concat} \left( \text{GAP}(F_p), \text{GMP}(F_p) \right) \right) \right),
\label{eq:pfm_weight}
\end{equation}

\begin{equation}
F_p' = F_p \times w_p.
\label{eq:pfm_output}
\end{equation}

In these equations, $F_p$ represents the input to the PFM module, and $F_p'$ is the output. $\text{GMP}(\cdot)$ denotes global max pooling, and $\text{Conv}_1$ represents the $7 \times 7$ convolutional layer. The output $F_p'$ is obtained by integrating both global and local perspectives through the attention mechanism, achieving the fusion of different views for refined feature representation.

\subsubsection{Class Discrimination Module }

To explicitly enhance the semantic separability of features with respect to object categories, we propose CDM which generates class-aware attention weights to emphasize category-relevant activations, as shown in Fig.~\ref{fig:structure}. This mechanism is particularly crucial in complex scenes with densely packed or visually ambiguous objects, where standard feature aggregation may dilute class-specific cues.
The class-aware attention generation process is illustrated in the following steps.
\begin{equation}
X_1 = \text{ReLU}(\text{BN}_1(\text{Conv}_1(F_c))) 
\label{eq:cdm_conv1}
\end{equation}

As shown in Equation~\eqref{eq:cdm_conv1}, the process begins by applying a $1 \times 1$ convolution to the input feature map $F_c$, followed by batch normalization and ReLU activation. This yields an intermediate representation $X_1$ that captures compact features with initial class-specific responses.
\begin{equation}
X_2 = \text{ReLU}(\text{BN}_2(\text{Conv}_2(X_1)))
\label{eq:cdm_conv2}
\end{equation}
This is then passed through a $3 \times 3$ convolution with batch normalization and ReLU, as formulated in Equation~\eqref{eq:cdm_conv2}, enabling $X_2$ to integrate richer contextual semantics that support more precise class-level discrimination.
\begin{equation}
w_c = \sigma(\text{Conv}_3(X_2)) 
\label{eq:cdm_attention_weight}
\end{equation}
Subsequently, a final $1 \times 1$ convolution and sigmoid activation are applied to $X_2$ in Equation~\eqref{eq:cdm_attention_weight} to generate the normalized attention weights $w_c$, which encode the relative importance of each feature stream with respect to class relevance.
\begin{equation}
F'_c = F_c \times w_c
\label{eq:cdm_output}
\end{equation}
As shown in Equation~\eqref{eq:cdm_output}, these weights are used to reweight the original input $F_c$ via element-wise multiplication, yielding the refined output $F'_c$ that selectively amplifies class-relevant information and enhances semantic separability among classes.

The notation follows the standard convention, where $F_c$ and $F'_c$ denote the input and output feature maps, respectively. $X_1$ and $X_2$ represent intermediate features within the attention generation process. $\text{Conv}_1$, $\text{Conv}_2$, and $\text{Conv}_3$ denote convolutional layers, with the first and third using $1 \times 1$ kernels for dimensionality adjustment, and the second using a $3 \times 3$ kernel to capture contextual information. $\text{BN}_1$ and $\text{BN}_2$ represent batch normalization layers applied after the first two convolutions. The output $F'\_c$ selectively emphasizes class-relevant features and improves class discrimination.

\section{Experiments}
In this section, we evaluate the proposed SPCI module through a series of experiments on both the NWPU VHR-10 dataset and the DIOR dataset. The NWPU VHR-10 benchmark is used to examine the module’s effectiveness in addressing key challenges in remote sensing object detection, such as small objects, complex backgrounds, and multi-scale features. To further assess generalization ability across diverse data domains, supplementary evaluations are conducted on the DIOR dataset. The experiments include detailed dataset descriptions, implementation settings, comparisons with state-of-the-art methods, and comprehensive ablation studies. The results consistently demonstrate the robustness and superiority of the SPCI module over baseline models.

\subsection{Dataset description}

In this study, we utilize two widely recognized remote sensing benchmarks to evaluate the performance and generalization ability of our proposed SPCI module: the NWPU VHR-10 dataset and the DIOR dataset.
NWPU VHR-10~\cite{NWPUVHR10} is a geospatial object detection dataset released by Northwestern Polytechnical University in 2014. It contains ten object categories, including airplanes, ships, storage tanks, baseball diamonds, tennis courts, basketball courts, ground track fields, harbors, bridges, and vehicles. The dataset consists of 800 high-resolution images, among which 650 images contain a total of 3,775 annotated objects, and 150 images are negative samples used for testing detection performance under challenging conditions. The spatial resolutions range from 0.5 to 2 meters for color images, and 0.08 meters for pan-sharpened color infrared images. The dataset presents challenges such as small object sizes, densely packed instances, and complex backgrounds, making it suitable for evaluating attention-based detection modules.

DIOR~\cite{dior2019} is a large-scale benchmark containing 23,463 images and over 190,000 annotated instances across 20 object categories, including harbors, airports, vehicles, bridges, and more. Each category appears in diverse environments and at various resolutions. Compared to NWPU VHR-10, DIOR introduces more semantic diversity and spatial complexity, enabling the evaluation of cross-dataset generalization. In this work, DIOR is used as a secondary testbed to assess the transferability of our model without any retraining or parameter adjustment.

\subsection{Experimental setup}

The experiments are conducted on a workstation equipped with an NVIDIA GeForce RTX 4060 GPU (8 GB VRAM), running PyTorch 2.0 in a Python 3.9 environment with CUDA 12.6. Supporting libraries include Albumentations 1.4.18 for data processing, OpenCV 4.10.0 for image handling, and Ultralytics 8.3.25 for YOLOv8-based implementations. These configurations ensure efficient training and evaluation.
The baseline model in this study is YOLOv8, which serves as the foundation for integrating the proposed SPCI module. SPCI is strategically inserted into the YOLOv8 backbone after the P3 and P5 stages. The first module enhances feature extraction for small object detection at the shallow level, while the second module refines multi-scale feature representation at a deeper level. These placements are designed to improve the model’s ability to detect small objects and effectively handle complex backgrounds by enhancing both local and global feature representation.

The modified YOLOv8 architecture achieves a balance between performance and computational efficiency, with 269 layers, 3.1M parameters, and a computational complexity of 8.3 GFLOPs. This lightweight design makes it suitable for high-resolution remote sensing datasets. The backbone outputs multi-scale features, which are passed to the Neck through operations such as upsampling and concatenation to facilitate multi-scale fusion. Finally, the detection head combines processed features from different scales to predict object bounding boxes and class probabilities.
Images are resized to 640×640 pixels and normalized to the \([0, 1]\) range, while annotations are reformatted into YOLO’s format, ensuring compatibility and efficient convergence during training. The training process begins with an initial learning rate of \(10^{-3}\), a batch size of 16, and utilizes the AdamW optimizer with a cosine annealing scheduler. These configurations enable robust feature learning over 500 epochs, balancing efficiency and convergence.

Performance evaluation is conducted using standard metrics, including precision (proportion of true positive predictions among all positives), recall (proportion of true positives detected), mAP50 (mean average precision at IoU=0.5), and mAP50-95 (mean average precision across IoU thresholds from 0.5 to 0.95). These metrics provide a comprehensive assessment of detection accuracy and robustness, particularly highlighting the model’s ability to address the challenges of small object detection and multi-scale representation.
All subsequent experiments, including those on the DIOR dataset, are conducted using this model as the baseline, with no further changes to the architecture, training configuration, or evaluation settings. This consistency ensures that performance differences observed across datasets are solely attributable to the experimental variables under investigation, allowing for fair cross-dataset evaluation.

\subsection{Comparison with state-of-the-art methods}

\begin{table*}[t] 
\centering
\caption{Comparison with state-of-the-art methods on the NWPU VHR-10 dataset. The best results are underlined.}
\resizebox{\textwidth}{!}{ 
\begin{tabular}{lccccccccccc}
\hline
\textbf{Method} & \textbf{mAP50 (\%)} & \textbf{Airplane} & \textbf{Ship} & \textbf{Storage Tank} & \textbf{Baseball Diamond} & \textbf{Tennis Court} & \textbf{Basketball Court} & \textbf{Ground Track Field} & \textbf{Harbor} & \textbf{Bridge} & \textbf{Vehicle} \\
\hline
Dynamic PL~\cite{wang2021dynamic} & 53.6 & 80.9 & 10.5 & 90.1 & 64.4 & 69.1 & 80.2 & 8.7 & 14.0 & 39.6 & 78.3 \\
SSD~\cite{liu2016ssd} & 76.0 & 95.7 & 83.1 & 85.9 & 96.7 & 82.3 & 85.9 & 57.9 & 54.6 & 42.1 & 75.4 \\
Deformable R-FCN~\cite{xu2017deformable} & 79.1 & 92.6 & 81.4 & 63.6 & 91.2 & 81.6 & 74.1 & 90.3 & 75.3 & 76.2 & 80.1 \\
Deformable DETR~\cite{zhu2021deformable} & 79.3 & 95.7 & 82.7 & 54.2 & 90.4 & 79.7 & 83.6 & 91.6 & 80.1 & 67.4 & 67.6 \\
RFN (SSD)~\cite{zhou2021rotated} & 80.6 & 92.2 & 63.8 & 79.3 & 91.2 & 83.1 & 82.7 & 91.6 & 80.1 & 78.5 & 60.7 \\
Faster R-CNN~\cite{ren2017faster} & 81.6 & 99.6 & 39.4 & 80.0 & 99.5 & 84.8 & 87.0 & \underline{100.0} & 87.3 & 81.2 & 57.4 \\
Soft-NMS~\cite{dong2019signms} & 82.3 & 88.3 & 80.3 & 60.1 & 90.4 & 80.6 & 89.7 & 98.9 & 90.1 & 67.3 & 77.6 \\
Sig-NMS~\cite{dong2019signms} & 82.9 & 90.8 & 80.5 & 59.2 & 90.8 & 80.8 & 90.9 & 99.8 & 90.3 & 67.8 & 78.1 \\
Faster R-CNN (VGG)~\cite{ren2017faster} & 84.0 & 99.7 & 88.6 & 39.7 & 90.9 & 80.3 & 97.0 & 99.7 & 88.9 & 75.3 & 74.4 \\
ARFP (YOLOv4)~\cite{zhang2021arfp} & 84.3 & 67.0 & 44.3 & 86.6 & 91.1 & 95.3 & 97.4 & 89.9 & 96.5 & 75.1 & \underline{99.5} \\
YOLOv5m~\cite{yolov5github} & 87.6 & 99.5 & 67.4 & 97.1 & 97.4 & 80.6 & 85.3 & \underline{100.0} & 80.5 & 80.8 & 87.4 \\
PR-Deformable DETR~\cite{chen2024pr} & 88.3 & 96.8 & 90.6 & 67.7 & 94.6 & 89.3 & 94.1 & \underline{100.0} & 90.1 & 80.5 & 80.0 \\
\textbf{Baseline (YOLOv8n)~\cite{yolov8}} & \textbf{88.9} & \textbf{95.1} & \textbf{92.0} & \textbf{93.3} & \textbf{91.8} & \textbf{88.9} & \textbf{86.9} & \textbf{97.9} & \textbf{\underline{99.1}} & \textbf{90.7} & \textbf{83.1} \\
YOLOv5l~\cite{yolov5github} & 89.1 & \underline{99.9} & 65.4 & 97.6 & 97.4 & 91.1 & 78.5 & \underline{100.0} & 83.6 & 86.9 & 90.4 \\
YOLOv7-tiny~\cite{wang2022yolov7} & 90.0 & 99.4 & \underline{95.8} & 58.7 & 96.9 & \underline{98.2} & 87.7 & 94.0 & 87.7 & \underline{99.5} & 81.7 \\
YOLOv5x~\cite{yolov5github} & 91.5 & \underline{99.9} & 69.8 & 98.0 & 97.5 & 93.5 & 90.9 & \underline{100.0} & 90.4 & 79.6 & 95.4 \\
\textbf{YOLO-SPCI (ours)} & \textbf{\underline{92.0}} & \textbf{95.1} & \textbf{94.1} & \textbf{\underline{99.5}} & \textbf{\underline{99.6}} & \textbf{86.4} & \textbf{\underline{100.0}} & \textbf{99.5} & \textbf{95.4} & \textbf{\underline{99.5}} & \textbf{87.7} \\
\hline
\end{tabular}
}
\label{tab:sota_comparison}
\end{table*}

\begin{table}[t]
\centering
\caption{Comparison with state-of-the-art methods on the DIOR dataset. The evaluation is conducted using mAP@50 and mAP@50–95 across 20 categories.}
\label{tab:dior_sota}
{\fontsize{9}{11}\selectfont
\setlength{\tabcolsep}{3.5pt}
\begin{tabular}{l|c|c}
\toprule
\textbf{Method}  & \textbf{mAP50 (\%)} & \textbf{mAP50–95 (\%)} \\
\midrule
Faster R-CNN~\cite{ren2015faster}         & 78.4 & -- \\
YOLOv5s~\cite{yolov5github}               & 79.5 & 59.3 \\
RetinaNet~\cite{lin2017focal}             & 77.1 & -- \\
Deformable DETR (small)~\cite{zhu2020deformable} & 80.0 & 60.1 \\
CenterNet~\cite{zhou2019objects}          & 75.3 & -- \\
YOLOv8-CTAM~\cite{yolov8ctam2024}         & 80.8 & 60.2 \\
YOLOv8-ESHA~\cite{yolov8esha2024}         & 80.9 & 60.5 \\
YOLOv8n (Baseline)~\cite{yolov8}          & 80.5 & 59.9 \\
\textbf{YOLO-SPCI} (Ours)                 & \textbf{81.2} & \textbf{60.9} \\
\bottomrule
\end{tabular}
}
\end{table}

To evaluate the effectiveness of the proposed SPCI module, we first compare it with a range of state-of-the-art object detection models on the NWPU VHR-10 dataset. Table~\ref{tab:sota_comparison} summarizes the results, highlighting the mAP50 and per-class performance metrics for each method.
As shown in Table~\ref{tab:sota_comparison}, YOLO-SPCI achieves the highest overall mAP50 score of 92.0\%, outperforming the baseline YOLOv8n (88.9\%) by a notable margin of 3.1\%. This improvement underscores the capability of the SPCI module in enhancing feature representation and optimizing multi-scale detection.

Specifically, YOLO-SPCI achieves the best performance in Storage Tank (99.5\%), Baseball Diamond (99.6\%), Basketball Court (100.0\%), and Bridge (99.5\%), ranking first in 4 out of the 10 categories. These results span a variety of challenging target types---from compact and densely distributed objects to large structures with complex geometries---demonstrating the adaptability of the proposed module to diverse spatial and semantic conditions.
Worth mentioning is that even in categories where YOLO-SPCI does not achieve the absolute best score, it still ranks among the top-performing models. For example, it achieves 95.4\% in Harbor and 87.7\% in Vehicle, maintaining strong performance under complex backgrounds and dense layouts, and reflecting consistent robustness and generalization ability.
Compared to more complex architectures such as PR-Deformable DETR (88.3\%) and YOLOv5x (91.5\%), YOLO-SPCI delivers a higher overall detection accuracy of 92.0\% while maintaining a lightweight design. This confirms the advantage of the proposed module in balancing precision and efficiency, making it well suited for practical deployment in remote sensing applications.

Beyond the NWPU VHR-10 dataset, we further evaluate the generalization capability of the proposed YOLO-SPCI framework on the large-scale DIOR benchmark.DIOR is a large-scale remote sensing benchmark that contains 20 object categories and over 23,000 annotated images, covering diverse scenes such as harbors, airfields, vehicles, and bridges.
Compared to NWPU VHR-10, the DIOR dataset offers a richer semantic spectrum and more complex spatial distributions, providing a challenging platform for cross-dataset validation. As reported in Table~\ref{tab:dior_sota}, YOLO-SPCI achieves the highest mAP50 (81.2\%) and mAP50-95 (60.9\%), outperforming both its baseline (YOLOv8n) and recent variants (CTAM, ESHA). Furthermore, while classical models like Faster R-CNN, RetinaNet, and CenterNet exhibit competitive results, their generalization is relatively weaker under the DIOR domain. Notably, Deformable DETR (small) reaches 80.0\% mAP50, still below our performance.
These results indicate that the proposed multi-dimensional attention module not only improves accuracy on the NWPU VHR-10 dataset, but also generalizes well to new data domains, highlighting its robustness and transferability.

\subsection{Ablation study}


\begin{table}[t]
\centering
\caption{Ablation study results on the NWPU VHR-10 dataset (\%)}
\label{tab:ablation_study}
{\fontsize{9}{11}\selectfont
\setlength{\tabcolsep}{3.5pt}
\begin{tabular}{clccccccc}
\toprule
\textbf{No.} & \textbf{Methods} & \textbf{SSG} & \textbf{PFM} & \textbf{CDM} & \textbf{Insertion} & \textbf{mAP50 (\%)} & \textbf{mAP50–95 (\%)} \\
\midrule
B1 & Baseline              & $\times$ & $\times$ & $\times$ & --          & 88.9 & 57.0 \\
B2 & Disable SSG           & $\times$ & $\checkmark$ & $\checkmark$ & P3+P5      & 90.4 & 58.8 \\
B3 & Disable PFM           & $\checkmark$ & $\times$ & $\checkmark$ & P3+P5      & 88.9 & 55.5 \\
B4 & Disable CDM           & $\checkmark$ & $\checkmark$ & $\times$ & P3+P5      & 91.2 & 58.0 \\
B5 & SPCI@P3 only          & $\checkmark$ & $\checkmark$ & $\checkmark$ & P3         & 90.3 & 56.2 \\
B6 & SPCI@P5 only          & $\checkmark$ & $\checkmark$ & $\checkmark$ & P5         & 90.7 & 56.7 \\
\textbf{B7} & \textbf{Ours (SPCI)} & \textbf{$\checkmark$} & \textbf{$\checkmark$} & \textbf{$\checkmark$} & \textbf{P3+P5} & \textbf{92.0} & \textbf{58.5} \\
\bottomrule
\end{tabular}
}
\end{table}

\begin{figure*}[t]
    \centering
    \includegraphics[width=\textwidth]{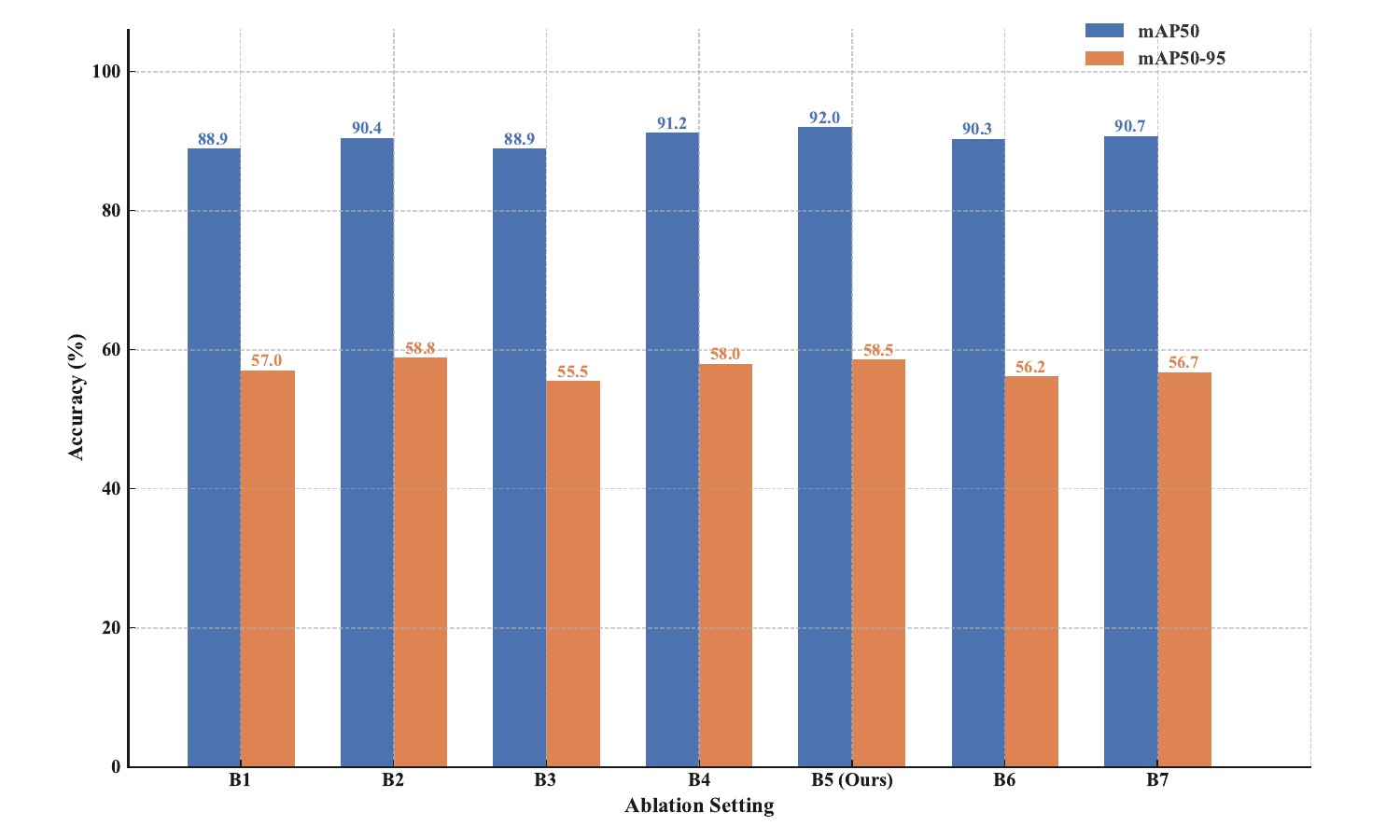}
    \caption{Quantitative comparison of mAP50 and mAP50-95 across all ablation variants (B1--B8).}
    \label{fig:ablation_bar}  
\end{figure*}

\begin{figure*}[t]
    \centering
    \includegraphics[width=\textwidth]{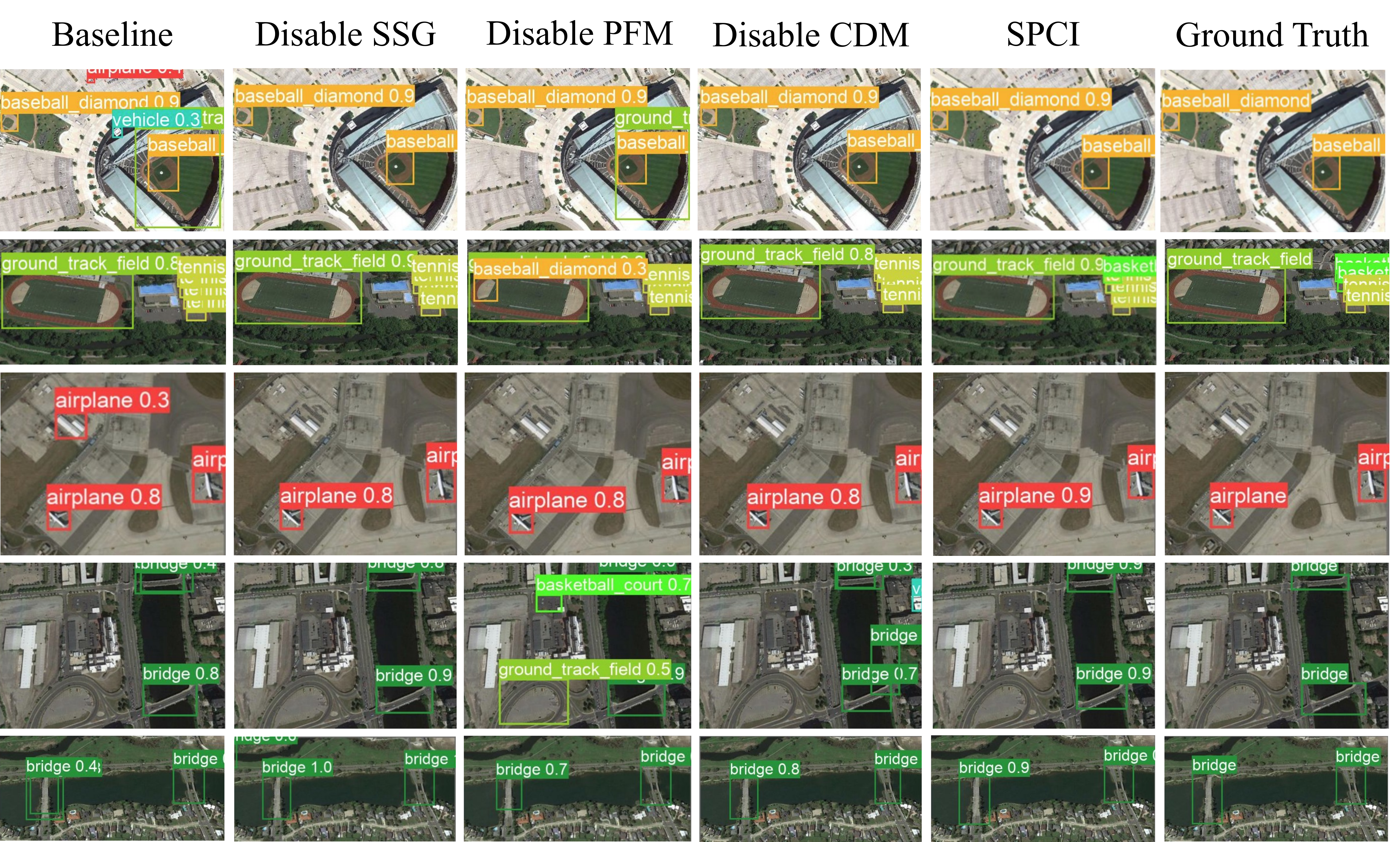} 
    \caption{Comparison of detection results between the baseline model and the proposed SPCI module with ablation studies on the NWPU VHR-10 dataset.}
    \label{fig:ablation_results}
\end{figure*}

To evaluate the individual contributions of each component in SPCI, an ablation study is conducted on the NWPU VHR-10 dataset. Specifically, we compare the detection performance of the baseline YOLOv8 model, models with disabled submodules (SSG, PFM, or CDM), and the complete SPCI module, as well as variants with different insertion strategies. The quantitative results are summarized in Table~\ref{tab:ablation_study}, while a bar chart in Fig.~\ref{fig:ablation_bar} provides an intuitive visual comparison across all B1–B8 configurations. Additionally, Fig.~\ref{fig:ablation_results} presents qualitative detection outcomes to highlight improvements in small object localization, background suppression, and inter-class differentiation.

We conduct comprehensive ablation experiments (B1–B7) to evaluate the contribution of individual components in the SPCI module and the impact of different insertion positions, as summarized in Table~\ref{tab:ablation_study}.
Starting from the baseline (B1), the complete model (B7) incorporating SSG, PFM, and CDM achieves the highest performance with an mAP50 of 92.0\% and mAP50-95 of 58.5\%. When SSG is removed (B2), the mAP50 drops to 90.4\%, showing that SSG plays a critical role in enhancing localization accuracy. Removing PFM (B3) causes the most significant performance degradation, with mAP50 falling to 88.9\% and mAP50-95 to 55.5\%, emphasizing its importance in spatial fusion and contextual feature integration. In contrast, disabling CDM (B4) results in a moderate reduction (mAP50: 91.2\%, mAP50-95: 58.0\%), demonstrating its role in refining class-specific features.

To investigate position sensitivity, we further evaluate three insertion strategies. B5 and B6 insert the complete SPCI module only at P3 and P5, respectively. Both yield improvements over the baseline (90.3\% and 90.7\% mAP50), yet still underperform the dual-position setting in B7. 

These trends confirm our design intuition. As discussed in Section Method, P3 corresponds to higher-resolution shallow features that benefit small object categories like airplane and basketball court, while P5 captures deeper semantic context, improving the detection of complex targets like bridge and harbor. The configuration in B7 (P3+P5 insertion) thus achieves optimal performance by leveraging complementary strengths across feature scales.

A visual summary of these results will be illustrated in Fig.~\ref{fig:ablation_bar}, providing an intuitive comparison across the B1–B7 variants. The bar chart highlights the performance gains brought by each module and insertion strategy. Notably, B7 (Ours), which integrates the full SPCI module at both P3 and P5 levels, exhibits the best overall performance in mAP50 and a strong result in mAP50-95. While B2–B4 help isolate the effect of disabling individual submodules, B5–B6 compare alternative placement strategies. B5 (SPCI@P3) and B6 (SPCI@P5) confirm that either shallow or deep placement alone improves over the baseline, but still underperform B7. 

These observations support our architectural design, where a balanced placement of the module at both shallow (P3) and deep (P5) layers enables the network to capture fine-grained localization cues and rich semantic context. As further demonstrated in Fig.~\ref{fig:ablation_results}, the benefits of this configuration are especially pronounced under challenging detection scenarios. It is worth noting that the qualitative comparisons focus on the baseline (B1), the three submodule-ablated variants (B2–B4), and the complete SPCI model (B7), while the position-based configurations (B5–B6) are included only in the quantitative evaluation presented in Fig.~\ref{fig:ablation_bar}.

Fig.~\ref{fig:ablation_results} illustrates the qualitative results across representative object categories, including baseball diamond, ground track field, tennis court, airplane, and bridge. For the baseball diamond category, the baseline and PFM-disabled models exhibit false positives, mistakenly detecting non-existent airplanes and vehicles. In contrast, the complete SPCI module accurately localizes the target region with a high confidence score (0.9), effectively suppressing false detections and aligning closely with the ground truth. For densely distributed categories such as ground track field, tennis court, and basketball court, the baseline and submodule-ablated models fail to detect all instances—especially in the case of basketball court—while only the full SPCI configuration successfully identifies all targets, highlighting its robustness in complex spatial layouts.

In small-object scenarios like airplane detection, the baseline produces false positives in background regions, and submodule-ablated models detect the correct target but with lower confidence (e.g., 0.8). The complete SPCI model, by contrast, achieves a more precise localization with higher confidence (0.9). For bridge detection in cluttered environments, the baseline generates redundant boxes with low confidence, and ablated models incorrectly detect unrelated classes such as basketball court, vehicle, and ground track field. Leveraging its integrated channel, spatial, and class-specific attention mechanisms, the complete SPCI model delivers accurate, high-confidence results closely aligned with the ground truth.

Taken together, the ablation results presented in Table~\ref{tab:ablation_study}, the comparative trends illustrated in Fig.~\ref{fig:ablation_bar}, and the qualitative detection performance in Fig.~\ref{fig:ablation_results} jointly demonstrate the effectiveness of the proposed SPCI module. By integrating SSG, PFM, and CDM into a unified architecture, the module successfully addresses limitations in feature selection, spatial fusion, and class-specific discrimination. The improvements are consistently reflected across quantitative metrics and visual precision, underscoring the robustness and generalizability of our design across diverse detection scenarios.

\section{Conclusion}
We presented YOLO-SPCI, a lightweight yet effective object detection framework for remote sensing imagery that integrates a Selective-Perspective-Class Integration (SPCI) module into the YOLOv8 backbone. By incorporating global stream regulation (SSG), perspective-aware fusion (PFM), and class-level discrimination (CDM), SPCI enhances multi-scale feature representation without altering the neck or head structure.
Extensive experiments on the NWPU VHR-10 dataset confirm that YOLO-SPCI outperforms several state-of-the-art methods in both accuracy and efficiency. These results highlight the generalizability and practicality of our multi-perspective integration strategy.
In future work, we plan to extend SPCI to transformer-based detectors and other sensing modalities such as multispectral or SAR data. Exploring deployment-friendly optimizations (e.g., pruning or quantization) and evaluating SPCI on larger datasets with greater category diversity are also promising directions.

{\footnotesize
\bibliographystyle{IEEEtran}
\bibliography{bibtex/bib/IEEEexample}
}

%




\end{document}